\documentclass[letterpaper]{article} 
\usepackage{aaai2026}  
\usepackage{times}  
\usepackage{helvet}  
\usepackage{courier}  
\usepackage[hyphens]{url}  
\usepackage{graphicx} 
\usepackage{natbib}  
\usepackage{caption} 
\usepackage{algorithm}
\usepackage{algorithmic}
\usepackage{xcolor}
\usepackage{multirow}
\usepackage{booktabs}
\usepackage{amssymb} 
\usepackage{bm}
\usepackage{amsfonts}
\usepackage{amsmath}
\usepackage[table]{xcolor}
\definecolor{Gray}{gray}{0.93}

\usepackage{newfloat}
\usepackage{listings}
\DeclareCaptionStyle{ruled}{labelfont=normalfont,labelsep=colon,strut=off} 
\floatstyle{ruled}
\newfloat{listing}{tb}{lst}{}
\floatname{listing}{Listing}
\title{Fine-Grained Representation for Lane Topology Reasoning}
\author{
    Guoqing Xu\textsuperscript{\rm 1,}\textsuperscript{\rm 2,}\equalcontrib ,
    Yiheng Li\textsuperscript{\rm 1,}\textsuperscript{\rm 2,}\equalcontrib\textsuperscript{\rm },
    Yang Yang \textsuperscript{\rm 1,}\textsuperscript{\rm 2,}\footnote{Corresponding author.}
}
\affiliations{

    \textsuperscript{\rm 1} MAIS, Institute of Automation, Chinese Academy of Sciences, Beijing, China\\
    \textsuperscript{\rm 2} School of Artificial Intelligence, University of Chinese Academy of Sciences, Beijing, China\\
    
    xuguoqing2024@ia.ac.cn, liyiheng2024@ia.ac.cn,
    yang.yang@nlpr.ia.ac.cn
}

\usepackage{bibentry}

\begin{document}

\maketitle

\begin{abstract}
Precise modeling of lane topology is essential for autonomous driving, as it directly impacts navigation and control decisions.
Existing methods typically represent each lane with a single query and infer topological connectivity based on the similarity between lane queries.
However, this kind of design struggles to accurately model complex lane structures, leading to unreliable topology prediction.
In this view, we propose a Fine-Grained lane topology reasoning framework (TopoFG).
It divides the procedure from bird’s-eye-view (BEV) features to topology prediction via fine-grained queries into three phases, i.e., Hierarchical Prior Extractor (HPE), Region-Focused Decoder (RFD), and Robust Boundary-Point Topology Reasoning (RBTR).
Specifically, HPE extracts global spatial priors from the BEV mask and local sequential priors from in-lane keypoint sequences to guide subsequent fine-grained query modeling.
RFD constructs fine-grained queries by integrating the spatial and sequential priors. 
It then samples reference points in RoI regions of the mask and applies cross-attention with BEV features to refine the query representations of each lane.
RBTR models lane connectivity based on boundary-point query features and further employs a topological denoising strategy to reduce matching ambiguity.
By integrating spatial and sequential priors into fine-grained queries and applying a denoising strategy to boundary-point topology reasoning, our method precisely models complex lane structures and delivers trustworthy topology predictions.
Extensive experiments on the OpenLane-V2 benchmark demonstrate that TopoFG achieves new state-of-the-art performance, with an OLS of 48.0\% on \textit{subset\_A} and 45.4\% on \textit{subset\_B}.
\end{abstract}

\begin{links}
    \link{Code}{https://github.com/GXmmm18/TopoFG}
\end{links}

\section{Introduction}
In autonomous driving, environmental perception is fundamental, covering tasks such as obstacle~\cite{li2025corenet,li2025rctrans} and lane detection~\cite{wang2023bev}. Building on lane perception, lane topology reasoning has recently gained attention for its ability to capture structural relationships between lanes and enable more reliable navigation in complex environments~\cite{he2024monocular}.
This not only involves the detection of lane centerlines and traffic elements, but also the reasoning of topological relationships such as lane connectivity and their associations with traffic elements.
Traditional methods often rely on cumbersome hand-crafted rules and post-processing to obtain relationships.
Recently, many methods tend to focus on using end-to-end approaches to unify the detection of lanes and traffic elements with the computation of their relationships into a single task, referred to as topology reasoning ~\cite{li2023graph}.

\begin{figure}[t]
    \centering
    \includegraphics[width=0.95\linewidth]{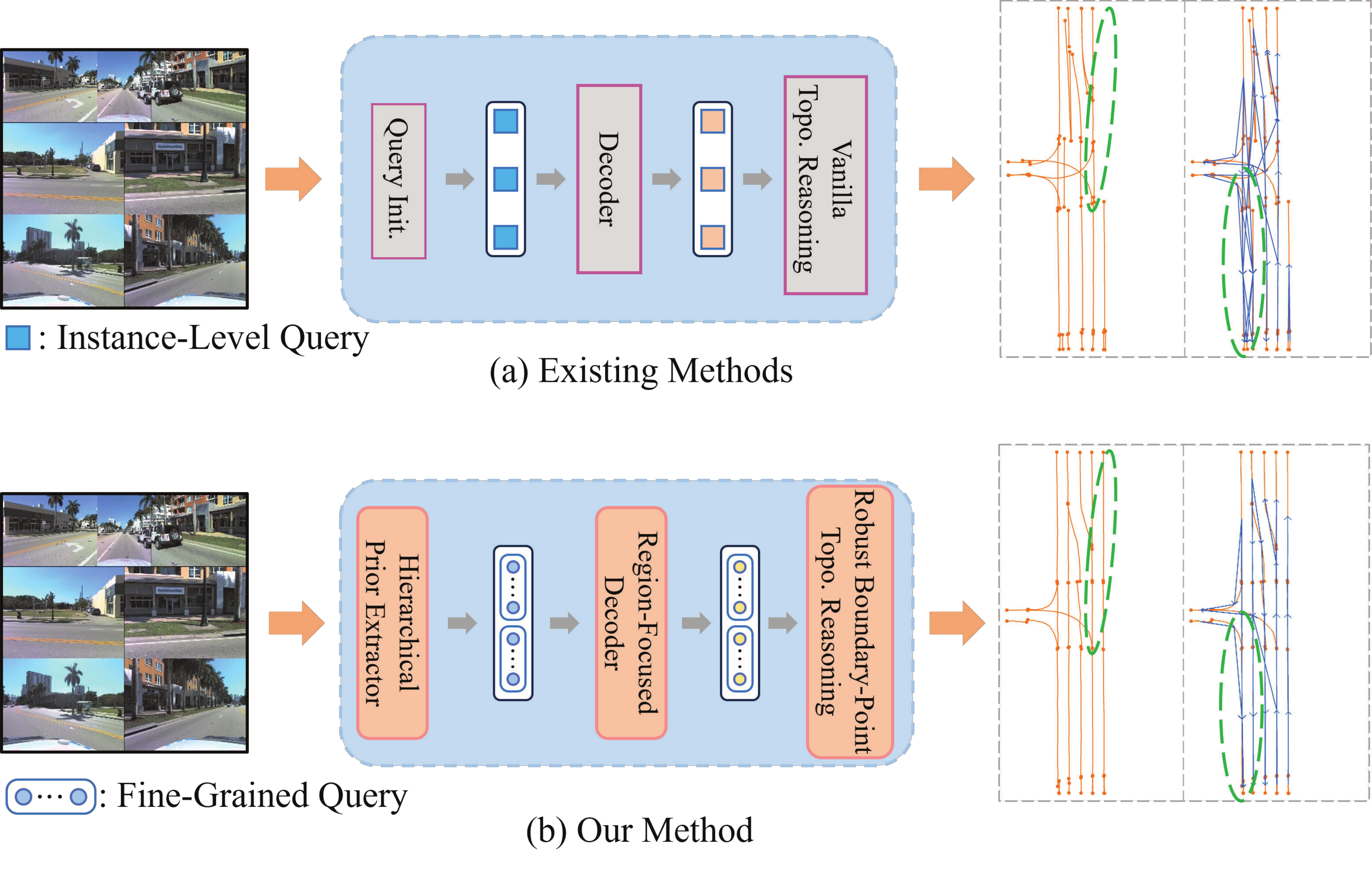}
    \caption{Comparison between existing methods and our method for lane topology reasoning. (a) \textbf{Existing Methods}: Using instance-level queries with coarse lane modeling and holistic topology reasoning, which may lead to incorrect predictions in complex scenes. (b) \textbf{Our Method}: Adopting fine-grained queries and boundary-point based topology reasoning for improved lane detection and topology prediction. The blue arrow on the right represents the lane topology connection, and the green dashed regions highlight the more reliable topology reasoning of our method.
    }
    \label{fig1}
\end{figure}

To implement end-to-end topology reasoning, recent studies have proposed unified frameworks that jointly detect lane centerlines and traffic elements while predicting their topological relationships. 
A common paradigm adopted by existing methods is to first detect individual lane centerlines and traffic elements, and then infer their topological relationships based on learned representations.
Approaches such as TopoNet~\cite{li2023graph} and TopoLogic~\cite{fu2024topologic} exemplify this framework, modeling each lane as a holistic entity and reasoning about their relationships at the instance level.
LaneSegNet\cite{li2023lanesegnet} enhances the semantic expressiveness of lane representations by modeling them as a series of lane segments.
Furthermore, it employs a lane attention mechanism to model lane segments.

Despite achieving promising performance, the existing works share a common limitation.
First, these approaches typically treat the entire centerline as a single unit and rely on a single query to predict all keypoints along the lane. 
This instance-level query modeling approach often lacks sufficient expressiveness when dealing with lanes that exhibit complex shapes or significant local variations. 
Furthermore, these approaches determine lane connectivity by directly computing the similarity between instance-level queries of different lanes. 
In practice, two lanes that are truly connected may only intersect locally at their boundary-points, making their overall similarity less obvious and thus harder to detect.
For example, consider three lanes where lane \(a\) ends ahead of two parallel lanes, \(b\) and \(c\), with similar geometries. While lane \(a\) should in fact lead to lane \(b\), instance-level query modeling may yield similar representations for \(b\) and \(c\), leading the model to incorrectly predict a connection from \(a\) to \(c\).

As shown in Figure~\ref{fig1}, we propose TopoFG.
In this approach, each lane is modeled as a sequence of fine-grained queries, with each query explicitly corresponding to a specific location along the lane.
This fine-grained representation allows the model to better capture local geometric variations and structural details.
Based on this representation, we perform topology reasoning using boundary-point queries and introduce a denoising strategy to improve prediction reliability.
We design the following three modules for the overall topology reasoning task to fully boost performance.
First, we introduce a Hierarchical Prior Extractor that derives global spatial priors from BEV masks and captures local sequential priors from in-lane sequences. 
Next, the Region-Focused Decoder integrates spatial and sequential priors into fine-grained queries and samples reference points from RoI regions of the mask, guiding the queries to focus on lane-relevant regions.
Finally, a Robust Boundary-Point Topology Reasoning module is introduced, which utilizes boundary-point queries, i.e., the start and end points of lanes, to infer topological connections and incorporates a denoising strategy to reduce ambiguity during training.

We conduct experiments on the widely used OpenLane-V2~\cite{wang2023openlane} dataset, and our method achieves superior performance by achieving 48.0\% OLS on \textit{subset\_A} and 45.4\% OLS on \textit{subset\_B}. 
Additionally, comprehensive ablation studies further demonstrate the effectiveness of each proposed module. Contributions are as follows:

\begin{itemize}
    \item We adopt fine-grained queries to represent a single lane, thereby enhancing the ability to model complex structures and improving the accuracy of topology prediction.
    \item We propose a Hierarchical Prior Extractor to help the model acquire prior information, a Region-Focused Decoder to reduce the interference from irrelevant information, and a Robust Boundary-Point Topology Reasoning to enhance the robustness of the reasoning results.
    \item We confirm the efficacy of the proposed method by achieving state-of-the-art performance on the OpenLane-V2 benchmark and superior accuracy and robustness in complex driving scenarios.
\end{itemize}

\section{Related Work}
\subsection{Lane Detection}
The primary objective of lane detection is to generate reliable geometric representations of lane boundaries or centerlines within the ego vehicle's surrounding environment, serving as critical input for downstream tasks like path planning and control.
Lane detection methods can generally be classified into 2D-based and 3D-based approaches.
2D lane detection typically includes segmentation-based~\cite{lee2017vpgnet,garnett20193d,abualsaud2021laneaf} and detection-based methods~\cite{tabelini2021polylanenet,han2022laneformer,wu2022yolop}.
Segmentation-based approaches perform pixel-wise prediction and distinguish lane instances using masks, grids, or keypoints, while detection-based approaches localize and classify lane instances in a single stage. 
These methods primarily operate in the image domain and are generally suited for relatively simple road scenarios~\cite{pan2018spatial,zhang2020ripple,qu2021focus}.
3D lane detection incorporates depth cues or geometric constraints to better reconstruct complex lane shapes~\cite{liu2022learning}. Existing methods can be grouped into BEV-based and BEV-free approaches. 
BEV-based methods~\cite{garnett20193d,pittner2024lanecpp} transform image features into BEV space to integrate height information for accurate localization, while BEV-free methods~\cite{tabelini2021keep,luo2023latr} either predict depth to project 2D lanes into 3D space or directly model lanes in 3D space.
Although significant progress has been made in the aforementioned methods, they still lack the ability to model road structures and connectivity, limiting their applicability in complex driving scenarios.

\begin{figure*}[t]
    \centering
    \includegraphics[width=0.9\linewidth]{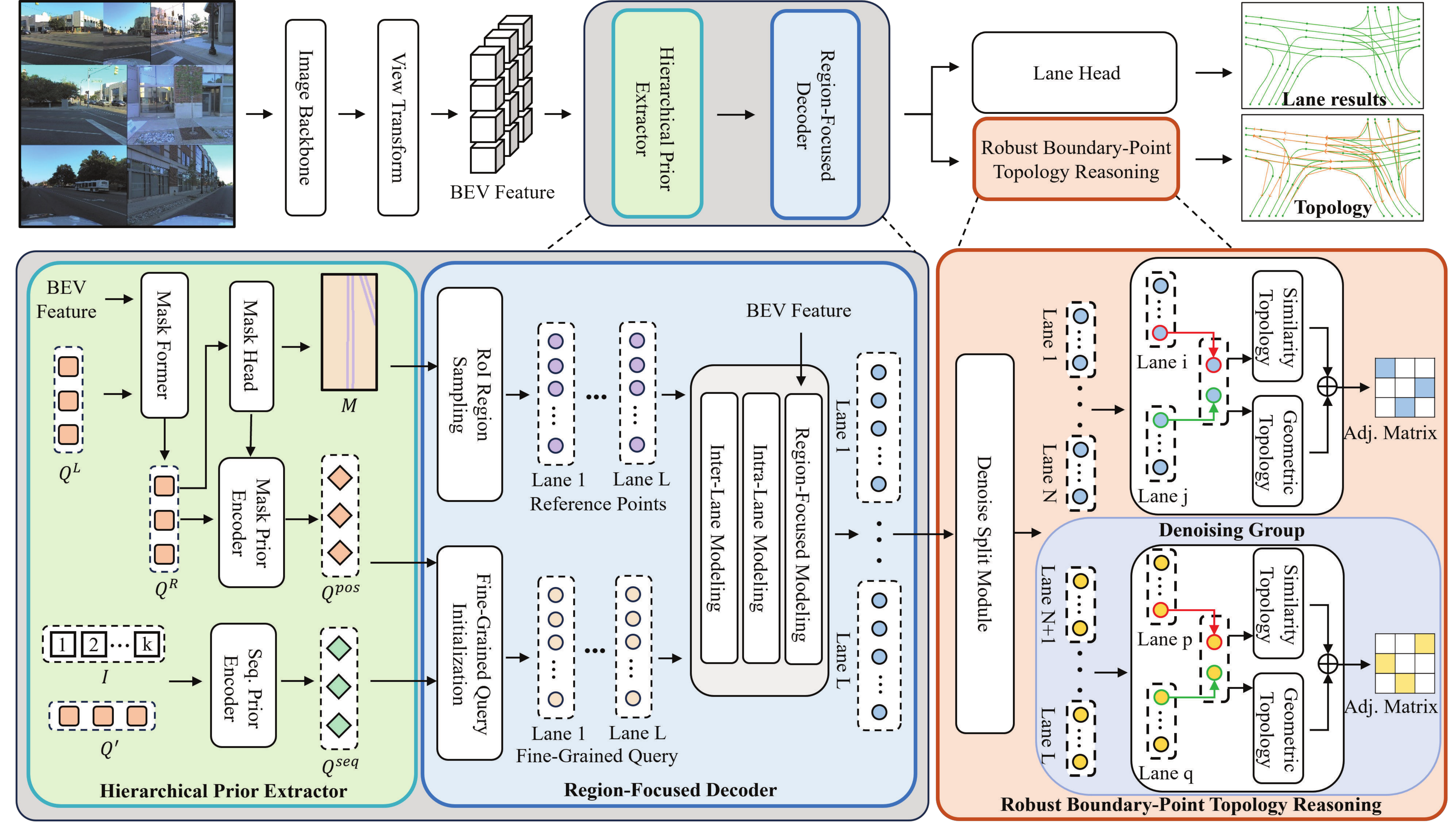}
    \caption{Overview of the TopoFG framework. The framework consists of three modules. First, the \textbf{Hierarchical Prior Extractor}, which extracts both spatial and local priors. Second, the \textbf{Region-Focused Decoder}, which enhances local geometric modeling by focusing on key lane regions. Third, the \textbf{Robust Boundary-Point Topology Reasoning} module, which constructs lane connectivity based on boundary-points and incorporates denoising training to improve structural stability.}  
    \label{overview}
\end{figure*}

\subsection{Vectorized HD Map}
The vectorized High-definition (HD) map method builds high-precision maps in real time using sensor data, providing a more flexible alternative to traditional HD maps.
HDMapNet~\cite{li2022hdmapnet} first generates BEV semantic segmentations through BEV feature extraction and a semantic segmentation model, followed by a time-consuming post-processing step to produce vectorized HD maps. 
VectorMapNet~\cite{liu2023vectormapnet} adopts the DETR~\cite{carion2020end} architecture, using transformer attention~\cite{vaswani2017attention} to decode the scene and directly predict ordered point sequences of map instances. 
MapTR~\cite{MapTR} represents map instances as ordered point sets and encodes them via hierarchical queries in a transformer decoder. 
It further introduces a permutation-equivalent modeling approach to resolve the order ambiguity in the matching stage. 
MapTRv2~\cite{liao2024maptrv2} extends this capability by incorporating deep supervision, enhanced feature encoding, and auxiliary losses to improve overall performance. 
Mask2Map~\cite{Mask2Map} first produces a rasterized map using a segmentation network, and then refines it into a vectorized map via a mask-driven network.
HD Map~\cite{zhou2024himap,li2024dtclmapper} has made progress in online lane reconstruction, but it is limited to lane reconstruction and lacks a deep understanding of the lane topology in the scene, making it difficult to effectively interpret the semantic information of the lane topology in the current scene.

\subsection{Lane Topology Reasoning}
Accurate lane topology modeling is critical for autonomous vehicles to comprehend drivable paths and traffic rules.
Previous methods have explored this problem from multiple perspectives.
STSU~\cite{can2021structured} detects lane centerlines from the BEV perspective and utilizes a Multi-Layer Perceptron (MLP) to predict connectivity between centerlines. 
In more complex road scenarios, TopoNet~\cite{li2023graph} treats lanes and traffic elements as nodes in a graph and employs Graph Neural Networks (GNNs)~\cite{zhou2020graph} to model the relationships between lanes and traffic signs.
SMERF~\cite{luo2024augmenting} incorporates Standard Definition Maps as additional information, enriching BEV feature representations and thereby improving the perception of lane centerlines. 
TopoMLP~\cite{wu2023topomlp}, based on PETR~\cite{liu2022petr}, performs lane centerline detection and employs a lightweight MLP to predict the topological relationships between centerlines.
LaneSegNet~\cite{li2023lanesegnet} models lanes as semantically enriched lane segments and introduces a Lane Attention mechanism to enhance the perception and feature representation of individual lane segments. 
TopoLogic~\cite{fu2024topologic} leverages the spatial positional relationships between lanes and proposes a concise and interpretable topology reasoning strategy, effectively modeling lane connectivity. 
Most of the previous works~\cite{can2022topology,can2023improving,liao2024lane} use coarse-grained lane modeling and predict the topological relationships between lanes by calculating the overall similarity of lanes. 
However, this design neglects the local geometric structure of lanes, making it difficult to effectively model complex lane shapes, resulting in unreliable topological relationship predictions.

In this paper, we propose TopoFG, which integrates prior information, fine-grained modeling, and boundary-point based topology reasoning.
TopoFG facilitates more reliable lane topology modeling and reasoning in dynamic driving scenes, ensuring accurate path planning and robust decision-making for vehicles in complex scenarios.

\section{Method}
\subsection{Overview}
As shown in Figure~\ref{overview}, we propose TopoFG, a novel framework designed for lane topology reasoning.
First, given multi-view camera images, we extract feature maps $\bm{F}_I \in \mathbb{R}^{C \times H \times W \times D}$ using a CNN-based backbone, where $C$, $H$, $W$, and $D$ denote the number of camera views, height, width, and number of channels of the image features, respectively.
Following ~\cite{li2024bevformer}, we adopt deformable attention~\cite{xia2022vision} to transform multi-scale image features into BEV features.
Inspired by previous works~\cite{Mask2Map}, we extend BEV features from a single scale to multiple scales.
The processed BEV features are then used as input to the Hierarchical Prior Extractor module.
In the Hierarchical Prior Extractor module, a mask prediction network is first employed to generate the BEV mask. We then separately extract the spatial prior from the BEV mask and the local sequential prior from lane point sequences.
Next, in the Region-Focused Decoder module, fine-grained queries are used to model lane geometry. With dynamic reference point sampling from RoI regions, the model focuses on key lane areas to enhance detection accuracy and geometric representation.
Finally, in the Robust Boundary-Point Topology Reasoning module, topology is modeled using start and end point queries. 
Boundary-point pairs are used to construct candidate connections, which are passed to a prediction head for Topology Reasoning. 
A denoising strategy is applied during training to guide the model toward more stable topological predictions.

\subsection{Hierarchical Prior Extractor}

To provide informative initialization for query embeddings, we design a Hierarchical Prior Extractor.
This module extracts two complementary priors, i.e., a global spatial prior from the predicted BEV mask and a local sequential prior that captures lane order.
We take the fused multi-scale BEV feature maps as input to this module.
Following Mask2Former~\cite{cheng2022masked}, we initialize a set of learnable queries \(\bm{Q}^L\), which interact with BEV features through the Mask Former module to produce refined queries \(\bm{Q}^R\).
These are then fed into the Mask Head to generate the final BEV lane masks \(\bm{M}\).
Based on the predicted BEV lane masks \(\bm{M}\), we compute a weight vector \(\bm{A}\).
The weight for each BEV location is calculated by applying a threshold \(\tau\) to the predicted probability:
\begin{equation}
\bm{A} = \bm{M} \cdot \mathbb{I}[\bm{M} \le \tau] 
+ \alpha \cdot \mathbb{I}[\bm{M} > \tau],
\end{equation}
where $\mathbb{I}[\cdot]$ is the indicator function, and $\alpha$ is a scaling factor that emphasizes high-confidence regions.
We adopt sine-cosine positional encoding to generate the spatial position encoding matrix \(\bm{P}\) for the BEV grid.
Given the weight vector \(\bm{A}\) and the refined queries \(\bm{Q}^R\), we derive the aggregated spatial prior queries \(\bm{Q}^{\text{pos}}\) through a weighted summation:

\begin{equation}
\bm{Q}^{\text{pos}} = \mathcal{N}_{\text{norm}}({\bm{P}^\top \cdot \bm{A}}) + \bm{Q}^R,
\end{equation}
where $\mathcal{N}_{\text{norm}}(\cdot)$ denotes a normalization operation.
  
To complement the spatial prior, we further introduce a local sequential prior that captures the inherent ordering and continuity of lane lines.
We first initialize a set of learnable queries \(\bm{Q}'\) to represent local queries along each lane line.
For each lane instance, we leverage the sequential nature of its constituent points by assigning them an ordered index.
By applying positional encoding followed by linear projection, the index sequence $I=\{1, \dots, k\}$ is transformed into an ordered embedding that preserves the local geometric structure.
The sequential prior $\bm{Q}^{\text{seq}}$ is computed as:
\begin{equation}
\bm{Q}^{\text{seq}} = \mathcal{F}(PE(I)) + \bm{Q}',
\end{equation}
where \(PE\) denotes a positional encoding function, e.g., sine-cosine or learnable, and $\mathcal{F}(\cdot)$ denotes an MLP.

\subsection{Region-Focused Decoder}

Accurately modeling lane topology demands capturing fine-grained local geometry. To this end, we introduce a Region-Focused Decoder, leveraging region-aware decoding for enhanced lane modeling.
During fine-grained query initialization, we incorporate both spatial and sequential priors into the fine-grained query features.
Given the spatial prior $\bm{Q}^{\text{pos}}$ of the \( i \)-th lane and the sequential prior \(\bm{Q}^{\text{seq}}\) of the  \( t \)-th keypoint, these two priors are then fused to produce the final query embedding as follows:
\begin{equation}
\bm{Q}^F_{i,t} = \bm{Q}^{pos}_i +  \mathcal{F}(\bm{Q}^{\text{seq}}_t)
\end{equation}
where \( \bm{Q}^F_{i,t} \) denotes the fine-grained query for the \( t \)-th keypoint of lane \( i \).

At each decoding layer \(m\), we adopt a two-stage self-attention mechanism for fine-grained queries.
We first perform inter-instance self-attention to capture interactions across different lane instances, followed by intra-instance self-attention to refine the point-level structure within each lane.
Subsequently, we adopt deformable attention~\cite{zhu2020deformable} as cross-attention to enable effective interaction between BEV features and queries.
To improve the spatial precision of query decoding, we adopt reference point sampling guided by the predicted lane mask, instead of using randomly initialized or fully learnable reference points \cite{fu2024topologic}.
These reference points \(\bm{R}\) constrain the attention to lane-relevant areas, thereby enhancing the model's ability to represent lane structures.
\begin{equation}
Q^{F, m+1}_{i,t} = \mathcal{A}_{\text{def}}\left(Q^{F, m}_{i,t}, \, \bm{F}_B, \, \bm{R}_{i,t} \right),
\end{equation}
where $\mathcal{A}_{\text{def}}(\cdot)$ denotes the Deformable Attention operation, and \(\bm{Q}_{i,t}^{F,m}\) denotes the \(t\)-th fine-grained query of the \(i\)-th lane at the \(m\)-th layer, \(\bm{F}_B\) represents the BEV feature, \(\bm{R}_{i,t}\) denotes the reference point coordinates associated with the \(t\)-th fine-grained query of the \(i\)-th lane.

\subsection{Robust Boundary-Point Topology Reasoning}

As the lane topology is determined by the relationships between endpoints, we retain only the first and last queries of each lane to serve as structural features for topology reasoning.
Specifically, each lane is represented by a sequence of fine-grained queries generated by the Region-Focused Decoder.
For lane $i$, the final representation consists of $k$ fine-grained queries $Lane_i=\{Q_{i,1}^F$, $Q_{i,2}^F$, $\dots$, $Q_{i,k}^F\}$. 
The first and last queries of each lane, \(Q_{i,1}^F\) and \(Q_{i,k}^F\), are selected as the boundary-point features of lane \(i\), and are denoted by \( f_{i}^{\text{start}} \) and \(f_{i}^{\text{end}}\), respectively, for subsequent topology reasoning.
Given any pair of lanes $(i, j)$, we concatenate the end-point feature of lane $i$ and the start-point feature of lane $j$, and feed this representation into a shared multilayer perceptron to predict connectivity:
\begin{align}
r_{i \rightarrow j} &= \text{Concat}(f_i^{end}, f_j^{start}) \\
\bm{S}_{i \rightarrow j} &= \mathcal{F}(r_{i \rightarrow j}),
\end{align}
where $\bm{S}_{i \rightarrow j}$ indicates whether lane $i$ leads to lane $j$, corresponding to the existence of a directed topological connection from lane $i$ to lane $j$.
We normalize the values of $\bm{S}$ using a Sigmoid function, which produces the Similarity Topology.
In parallel, we compute the Euclidean distance between the boundary-points of all lane pairs to obtain a geometric distance matrix, which is then mapped to a Geometric Topology using a mapping function following the~\cite{fu2024topologic} design.
Finally, the Similarity Topology and Geometric Topology are summed to obtain the final adjacency matrix for lane topology reasoning.

The number of predicted lanes often exceeds that of ground-truth lanes, necessitating Hungarian matching~\cite{kuhn1955hungarian} to compute one-to-one correspondences for loss calculation. 
Based on these matches, the ground-truth adjacency matrix is mapped into a larger zero-initialized matrix to serve as supervision. 
However, this process depends heavily on the matching results, causing the supervision matrix for the same sample to vary across epochs. 
Such instability degrades the model’s ability to learn topological relationships and hinders effective topology reasoning.

To improve the robustness and reliability of supervision in topology modeling, we propose a denoising training mechanism guided by boundary points. 
The key idea is to generate noisy queries from each ground-truth (GT) instance, allowing the supervision adjacency matrix to be fixed in advance and providing stable learning signals.
Given \(N_{gt}\) GT instances and \(G\) denoising groups, we construct \(N_{gt} \times G\) noisy queries, which are refined along with vanilla queries by the model. 
The refined queries are then split into vanilla and denoising queries, each used for separate topology reasoning.
For denoising queries, the original adjacency matrix is expanded \(G\) times into a block-diagonal form, which serves as supervision during training to enhance the model’s robustness and structural consistency.
During inference, only vanilla queries are used for topology reasoning, and denoising queries are discarded.
This strategy provides stable topology labels, thereby improving the accuracy and robustness of topology reasoning.

\subsection{Lane Head}
After refinement by the decoder, the queries are decoded by the lane head into lane centerline coordinates. 
Similar to topology reasoning, the denoising queries are discarded during inference.
We follow the ~\cite{fu2024topologic}, with the only difference being that each query is responsible for predicting a single key point on the lane.

\section{Experiments}

\subsection{Datasets and metric}

\textbf{Dataset}. We conduct experiments on the OpenLane-V2 benchmark ~\cite{wang2023openlane}. 
Designed specifically for autonomous driving scenarios, OpenLane-V2 is a large-scale multi-task benchmark that covers lane centerline detection, traffic element detection, and topology reasoning tasks. 
Built upon Argoverse2~\cite{wilson2023argoverse} and nuScenes~\cite{caesar2020nuscenes}, the dataset consists of two subsets: \textit{subset\_A} and \textit{subset\_B}, each containing 1,000 scenes. 
All data are collected at 2Hz with multi-view images. 
These annotations include lane centerlines, 
traffic elements in the front-view, 
as well as topological relations between lanes and between lanes and traffic elements. 
\textit{subset\_A} and \textit{subset\_B} contain 7 and 6 camera views, respectively.

\textbf{Metric}.
We follow the official evaluation protocol of OpenLane-V2 to assess both perception and topology reasoning capabilities. 
The evaluation metrics include DET$_{l}$, DET$_{t}$, TOP$_{ll}$, and TOP$_{lt}$. 
Specifically, DET$_{l}$ measures alignment between predicted and ground‑truth lane centerlines using the Fréchet distance.
DET$_{t}$ measures traffic‑element detection quality using Intersection over Union (IoU).
TOP$_{ll}$ and TOP$_{lt}$ evaluate the structural correctness of lane-lane and lane–traffic element connections, respectively. 
The OpenLane-V2 Score (OLS) provides a unified evaluation by aggregating metrics from various perspectives of the primary task:
\begin{equation}
\text{OLS} = \frac{1}{4} \left[ \text{DET}_l + \text{DET}_t + \sqrt{\text{TOP}_{ll}} + \sqrt{\text{TOP}_{lt}} \right].
\end{equation}
All TopoFG evaluations are conducted using the metrics defined in the latest OpenLane-V2 v2.1.0 release.

\subsection{Implementation Details}

\begin{table*}[ht]
\centering
\renewcommand{\arraystretch}{1} 
\begin{tabular}{ll|lccccc}
\bottomrule
\hline \noalign{\smallskip}
{Dataset} & {Method} & {Venue} & {OLS$\uparrow$} & {DET$_l$ $\uparrow$}  & {DET$_t$ $\uparrow$} & {TOP$_{ll}$ $\uparrow$} & {TOP$_{lt}$ $\uparrow$}  \\ \noalign{\smallskip} \hline \noalign{\smallskip}
\multirow{7}{*}{ \textit{subset\_A}} & STSU~\cite{can2021structured} & ICCV2021 & 29.3 & 12.7 & 43.0 & 2.9 & 19.8 \\
        & VectorMapNet~\cite{liu2023vectormapnet} & ICML2023 & 24.9 & 11.1 & 41.7 & 2.7 & 9.2 \\
        & MapTR~\cite{MapTR} & ICLR2023 & 31.0 & 17.7 & 43.5 & 5.9 & 15.1\\
        & TopoNet~\cite{li2023graph} & Arxiv2023 & 39.8    & 28.6 & \underline{48.6} & 10.9 & 23.8 \\
        & TopoMLP~\cite{wu2023topomlp}$^{\dag}$ & ICLR2024 & \underline{44.1} & 28.5 & {\bf49.5} & 21.7 & \underline{26.9}  \\
        & TopoLogic~\cite{fu2024topologic} & NeurIPS2024 & \underline{44.1} & \underline{29.9} & {47.2}   & {\underline{23.9}} & 25.4 \\
        \rowcolor{Gray} & TopoFG(Ours) & - & \bf 48.0(+3.9) & \bf 33.8(+3.9) & {47.2}   & {\bf 30.8(+6.9)} & \bf 30.9(+4.0) \\   
\bottomrule \noalign{\smallskip}
                           
\multirow{6}{*}{ \textit{subset\_B}}  & STSU~\cite{can2021structured} & ICCV2021 & - & 8.2 & 43.9 & - & - \\
        & VectorMapNet~\cite{liu2023vectormapnet} & ICML2023 & - & 3.5 & 49.1 & - & - \\
        & MapTR~\cite{MapTR} & ICLR2023 & - & 15.2 & 54.0 & - & - \\
        & TopoNet~\cite{li2023graph} & Arxiv2023 & 36.8 & 24.3   & \textbf{55.0}     & 6.7     & 16.7    \\
        & TopoLogic~\cite{fu2024topologic} & NeurIPS2024                      & \underline{42.3}    & \underline{25.9}   & \underline{54.7}  & \underline{21.6}    & \underline{17.9}   \\
        \rowcolor{Gray} & TopoFG(Ours)  & - & {\bf 45.4(+3.1)}  & {\bf 30.0(+4.1)}      & { 53.0}      & {\bf 27.2(+5.6)} &{\bf 21.7(+3.8)} 
    \\  \bottomrule \hline \noalign{\smallskip}
\end{tabular}
\caption{Comparisons of our model and existing state-of-the-art methods on \textit{subset\_A} and  \textit{subset\_B}.
``-'' denotes the absence of relevant data. The best performances are highlighted in {\bf bold}, while the second one is \underline{underlined}. ``$+$" indicates the absolute improvements compared with the second one. ``${\dag}$" indicates the result evaluated using the officially released model. }
\label{main_table}
\end{table*}

For image inputs, the resolution is set to $800 \times 1024$ for \textit{subset\_A} and $480 \times 800$ for \textit{subset\_B}. 
A ResNet-50~\cite{he2016deep} backbone is used to extract features from surround-view images, followed by a Feature Pyramid Network~\cite{lin2017feature} to generate multi-scale representations.    
In the lane centerline detection task, the model predicts 200 lanes, with each lane represented by $k = 11$ fine-grained queries.
A decoder with $m=6$ layers is used to refine the queries.
We use $G=5$ denoising groups, each containing \(N_{gt}\) queries. 
For the Hierarchical Prior Extractor, we set the threshold \(\tau\) to 0.3 and the scaling factor \(\alpha\) to 1.0. 
The AdamW optimizer is adopted with a weight decay of 0.01 and an initial learning rate of $2 \times 10^{-4}$. 
A cosine annealing strategy is used for learning rate scheduling, with 500 warmup steps.
All experiments are conducted on 8 NVIDIA A100 GPUs for 24 epochs, with a batch size of 1 per GPU.

\subsection{Performance Comparison}
 
As shown in Table \ref{main_table}, the main results on \textit{subset\_A} and \textit{subset\_B} are presented.
We compare the proposed TopoFG with existing state-of-the-art methods on the OpenLane-V2 dataset. 
The results show that TopoFG consistently achieves superior performance across both subsets. 
On \textit{subset\_A}, TopoFG achieves an OLS score of 48.0\%, significantly outperforming TopoLogic (44.1\%) and TopoMLP (44.1\%). 
For lane centerline detection, it attains 33.8\% on $\text{DET}_{l}$. 
In terms of topology reasoning, it achieves 30.8\% on $\text{TOP}_{ll}$ and 30.9\% on $\text{TOP}_{lt}$, indicating stronger structural modeling capability compared to existing methods. 
On \textit{subset\_B}, TopoFG also achieves the best overall performance with an OLS of 45.4\%. 
It obtains 27.2\% on $\text{TOP}_{ll}$ and 21.7\% on $\text{TOP}_{lt}$, demonstrating stable and generalizable topology reasoning in complex scenarios.
As shown in Table \ref{laneseg}, the performance of TopoFG is compared with existing methods on the OpenLane-V2 lane segment detection benchmark.
TopoFG outperforms all other methods in all metrics, achieving an mAP score of 34.4\%, surpassing TopoLogic and Topo2Seq~\cite{yang2025topo2seq}. Additionally, TopoFG excels in topology reasoning compared to other methods.

\begin{table}[t]
    \centering
    \resizebox{1\linewidth}{!}{
    \begin{tabular}{lccccc}
    \toprule
       Method \ & mAP$\uparrow$  & AP$_{ls}\uparrow$ & AP$_{ped}\uparrow$ & TOP$_{ls}\uparrow$ \\
         \midrule
         TopoNet 
         & 23.0 & 23.9    &  22.0 & -   \\
         MapTR 
         & 27.0 & 25.9    &  28.1  & -   \\
         MapTRv2 
         & 28.5 & 26.6    &  30.4  & -    \\
         LaneSegNet
         & 32.6 & 32.3    &  32.9  & 25.4 \\
         TopoLogic
         & 33.2 & 33.0    &  33.4  & \underline{30.8}   \\
         Topo2Seq
         & \underline{33.6} & \underline{33.7}    &  \underline{33.5}  & 26.9   \\
         \rowcolor{Gray} TopoFG (Ours)    & \bf{34.4(+0.8)} & \bf{33.8(+0.1)}    &  \bf{35.1(+1.6)} & \bf{31.3(+0.5)}   \\
    \bottomrule      
    \end{tabular}
    }
    \caption{Performance comparison with state-of-the-art methods on OpenLane-V2 benchmark on lane segment. 
    ``-'' denotes the absence of relevant data. The best performances are highlighted in {\bf bold}, while the second one is \underline{underlined}. ``$+$" indicates the absolute improvements compared with the second one.}
    \label{laneseg}
\end{table}

\begin{figure*}[t]
    \centering
    \includegraphics[width=0.95\linewidth]{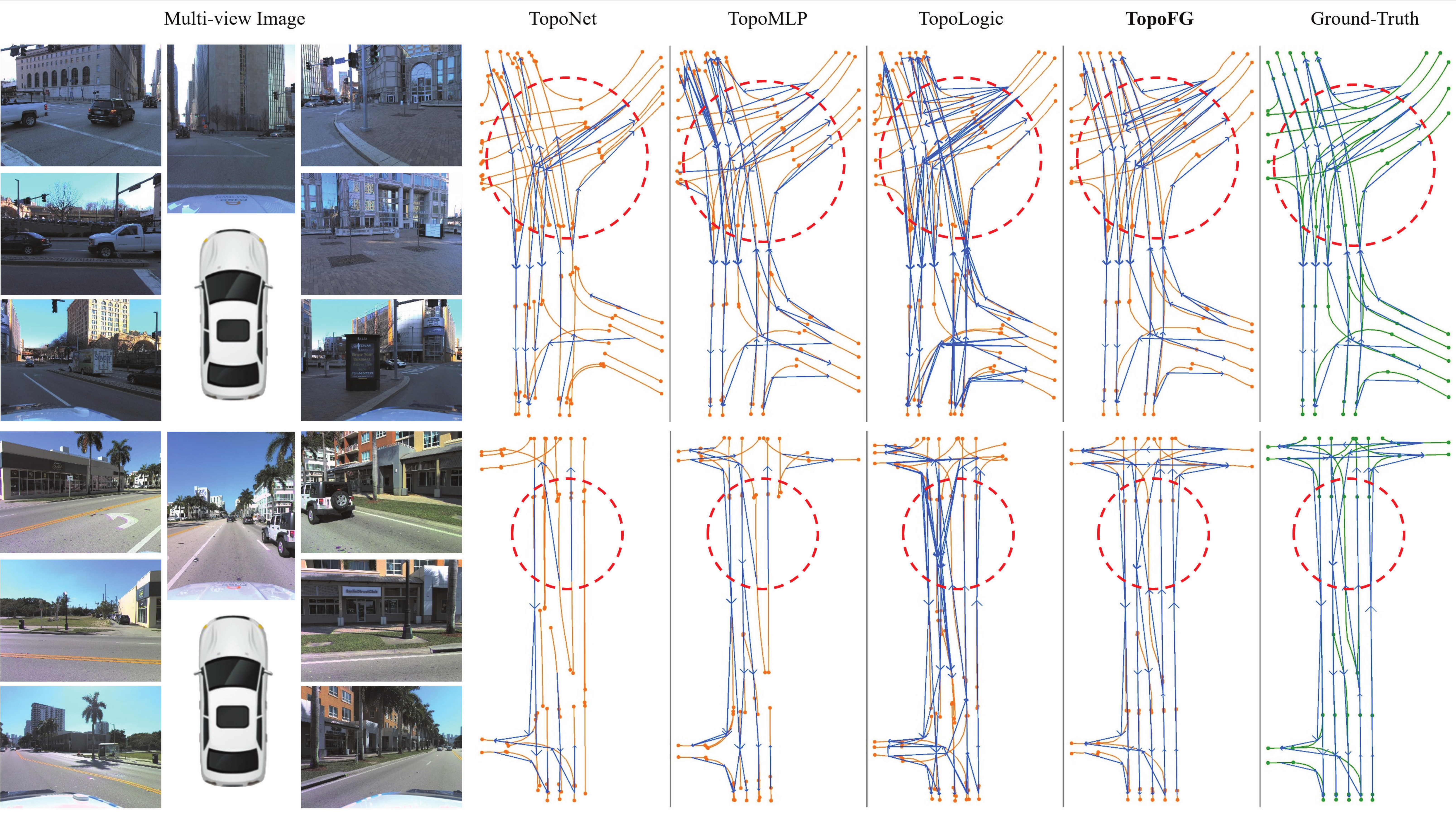}
    \caption{Qualitative comparison of different methods on the lane topology reasoning task. From left to right: multi-view input images, TopoNet, TopoMLP, TopoLogic, our proposed TopoFG, and the ground truth. The figure shows the predicted lane centerlines in the bird’s-eye view, where orange lines represent the predicted lane centerlines and blue arrows indicate the topological connections between lanes.} 
    \label{Visualization}
\end{figure*}

\subsection{Ablation Study}
\textbf{Contributions of Main Components.}
To thoroughly assess the contribution of each component in TopoFG, we carry out ablation experiments on the OpenLane-V2 \textit{subset\_A}. 
As shown in Table~\ref{main_ablation}, each module in TopoFG consistently improves performance. 
Compared to the baseline~\cite{fu2024topologic}, our proposed HPE achieves significant improvements in OLS and DET$_{l}$, while RFD further enhances both detection and topology-related metrics. 
RBTR yields the best overall performance, boosting OLS to 48.0\% and significantly improving TOP$_{ll}$ and TOP$_{lt}$, demonstrating the effectiveness of TopoFG in topology reasoning.
\begin{table}[t]
    \centering
    \resizebox{0.48\textwidth}{!}{
    \begin{tabular}{lccccc}
    \toprule
        \ & {OLS$\uparrow$} & {DET$_l$ $\uparrow$}  & {DET$_t$ $\uparrow$} & {TOP$_{ll}$ $\uparrow$} & {TOP$_{lt}$ $\uparrow$}  \\
         \midrule
         baseline
        & 44.1 & 29.9 & 47.2   & 23.9 & 25.4 \\
         +HPE   
         & 45.4 & 31.3 & 47.5 & 26.1 & 26.6  \\
         +RFD   
         & 45.8 & 31.8 & 47.2 & 26.8 & 27.7  \\
         \rowcolor{Gray} +RBTR
         & 48.0 & 33.8 & 47.2 & 30.8 & 30.9 \\
    \bottomrule      
    \end{tabular}
    }
    \caption{Contributions of Main Components. The main components include the Hierarchical Prior Extractor (HPE), Region-Focused Decoder (RFD), and Robust Boundary-Point Topology Reasoning (RBTR). Each module brings consistent performance gains.}
    \label{main_ablation}
\end{table}

\textbf{Contributions of Submodules.}
As shown in Table~\ref{HPE_test}, we progressively introduce the local sequential prior and global spatial prior to evaluate their effects. 
The introduction of local sequential prior leads to notable improvements in OLS and DET$_{l}$, and further adding global spatial prior continuously enhances all metrics, with OLS increasing to 45.4\%. 
These results indicate that the two priors are complementary and jointly contribute to performance gains.
\begin{table}[t]
    \centering
    \resizebox{0.48\textwidth}{!}{
    \begin{tabular}{cc|ccccc}
    \toprule
        \ LP & GP & {OLS$\uparrow$} & {DET$_l$ $\uparrow$}  & {DET$_t$ $\uparrow$} & {TOP$_{ll}$ $\uparrow$} & {TOP$_{lt}$ $\uparrow$}  \\ 
         \midrule
        &   & 44.1 & 29.9 & 47.2   & 23.9 & 25.4 \\
        \checkmark &   & 45.1 & 31.2 & 45.5 & 26.7 & 27.3  \\
         & \checkmark & 45.2 & 31.0 & 45.4 & 26.7 & 28.0  \\
         \rowcolor{Gray} \checkmark & \checkmark & 45.4 & 31.3 & 47.5 & 26.1 & 26.6 \\
    \bottomrule      
    \end{tabular}
    }
    \caption{Ablation study for evaluating the contributions of the Hierarchical Prior Extractor, including the local sequential prior (LP) and global spatial prior (GP).}
    \label{HPE_test}
\end{table}
As shown in Table~\ref{RFD_test}, fine-grained query initialization effectively enhances lane representation capability. 
When combined with sampled reference points, the model achieves the best performance across all metrics. 
These results indicate that the design improves both the accuracy of lane modeling and the perception of key regions.
\begin{table}[t]
    \centering
    \resizebox{0.48\textwidth}{!}{
    \begin{tabular}{cc|ccccc}
    \toprule
        \ FQI & SRP & {OLS$\uparrow$} & {DET$_l$ $\uparrow$}  & {DET$_t$ $\uparrow$} & {TOP$_{ll}$ $\uparrow$} & {TOP$_{lt}$ $\uparrow$}  \\ 
         \midrule
        &   & 45.4 & 31.3 & 47.5 & 26.1 & 26.6 \\
        \checkmark &   & 45.6 & 31.0 & 45.8 & 28.6 & 27.2  \\
        \rowcolor{Gray}  \checkmark & \checkmark & 45.8 & 31.8 & 47.2 & 26.8 & 27.7 \\
    \bottomrule      
    \end{tabular}
    }
    \caption{Ablation study for evaluating the contributions of the Region-Focused Decoder, including fine-grained query initialization (FQI) and sampled reference points (SRP).}
    \label{RFD_test}
\end{table}
\begin{table}[t]
    \centering
    \resizebox{0.48\textwidth}{!}{
    \begin{tabular}{cc|ccccc}
    \toprule
        \ BTR & DTR & {OLS$\uparrow$} & {DET$_l$ $\uparrow$}  & {DET$_t$ $\uparrow$} & {TOP$_{ll}$ $\uparrow$} & {TOP$_{lt}$ $\uparrow$}  \\ 
         \midrule
        &    & 45.8 & 31.8 & 47.2 & 26.8 & 27.7 \\
        \checkmark &   & 46.6 & 33.0 & 45.8 & 28.3 & 29.6  \\
         & \checkmark & 47.3 & 33.4 & 46.4 & 29.0 & 30.7 \\
         \rowcolor{Gray} \checkmark & \checkmark & 48.0 & 33.8 & 47.2 & 30.8 & 30.9 \\
    \bottomrule      
    \end{tabular}
    }
    \caption{Ablation study to assess the effectiveness of Robust Boundary-Point Topology Reasoning, which is based on boundary-point topology reasoning (BTR) and denoised topology reasoning (DTR).}
    \label{RBTR_test}
\end{table}
As shown in Table~\ref{RBTR_test}, the boundary-point topology reasoning effectively enhances the recognition of topological connections.
When combined with denoised topology reasoning, the overall performance is further improved, with OLS reaching 48.0\%, while TOP$_{ll}$ and TOP$_{lt}$ increase to 30.8\% and 30.9\%, respectively.
These results demonstrate the robustness and reliability of the proposed method in performing topology reasoning.

\subsection{Qualitative Analysis}
We compare several mainstream methods on the lane topology reasoning task, as shown in Figure \ref{Visualization}. 
From left to right, the results include multi-view input images, TopoNet~\cite{li2023graph}, TopoMLP~\cite{wu2023topomlp}, TopoLogic~\cite{fu2024topologic}, our TopoFG, and the ground truth. 
All predictions are visualized in a bird’s-eye view. 
Orange lines represent the predicted lane centerlines, and blue arrows indicate the topological connections between lanes.
It can be observed that TopoNet, TopoMLP, and TopoLogic suffer from missing lanes and incorrect connections, particularly at intersections. 
In contrast, TopoFG better recovers complete lane structures and accurately captures topological relationships.

\section{Conclusion}
In this work, we present an end-to-end fine-grained lane topology reasoning framework that addresses the limitations of previous methods relying on a single query to represent each lane.
By representing each lane with multiple spatially-aware queries, our proposed method captures local geometric variations and enables more accurate topological inference. 
The proposed denoised topology reasoning strategy effectively improves the robustness and reliability of learning topological relationships between lanes.
Extensive experiments on the OpenLane-V2 benchmark validate the effectiveness of our method, which achieves 48.0\% OLS on \textit{subset\_A} and 45.4\% OLS on \textit{subset\_B}, outperforming previous state-of-the-art methods in topology reasoning.

\section{Acknowledgments}
This work was supported in part by Beijing Natural Science Foundation under no. L221013  and Chinese National Natural Science Foundation Projects 62206276.
\bibliography{aaai2026}

\end{document}